# GeneAgent: Self-verification Language Agent for Gene Set Knowledge Discovery using Domain Databases

Zhizheng Wang[a, ǂ], Qiao Jin[a, ǂ], Chih-Hsuan Wei[a], Shubo Tian[a], Po-Ting Lai[a], Qingqing Zhu[a], Chi-Ping Day[b], Christina Ross[b], Zhiyong Lu[a, *]

[a] National Center of Biotechnology Information (NCBI), National Library of Medicine (NLM), National Institutes of Health (NIH), Bethesda, MD 20894, USA

[b] Laboratory of Cancer Biology and Genetics, Center for Cancer Research, National Cancer Institute (NCI), National Institutes of Health (NIH), Bethesda, MD 20894, USA

ǂ Authors contributed equally

* Correspondence: zhiyong.lu@nih.gov

## Abstract

Gene set knowledge discovery is essential for advancing human functional genomics. Recent studies have shown promising performance by harnessing the power of Large Language Models (LLMs) on this task. Nonetheless, their results are subject to several limitations common in LLMs such as hallucinations. In response, we present GeneAgent, a first-of-its-kind language agent featuring self-verification capability. It autonomously interacts with various biological databases and leverages relevant domain knowledge to improve accuracy and reduce hallucination occurrences. Benchmarking on 1,106 gene sets from different sources, GeneAgent consistently outperforms standard GPT-4 by a significant margin. Moreover, a detailed manual review confirms the effectiveness of the self-verification module in minimizing hallucinations and generating more reliable analytical narratives. To demonstrate its practical utility, we apply GeneAgent to seven novel gene sets derived from mouse B2905 melanoma cell lines, with expert evaluations showing that GeneAgent offers novel insights into gene functions and subsequently expedites knowledge discovery.

## Introduction

Genomics has been a research interest of molecular biologists for a long time[1,2,3,4]. Numerous mRNA expression experiments and proteomics investigations have yielded sets of genes and proteins that may be differentially expressed or co-modified[5,6]. In these cases, the fundamental premise is that the identified genes in a set should be involved in the most relevant biological processes or molecular functions. Therefore, it becomes imperative to elucidate the mechanisms underpinning co-abundance and physical interactions among multiple genes.

Gene set enrichment analysis (GSEA), as the representative research in genomics, aims to measure the over-representation or under-representation of biological functions associated with a set of genes or proteins[7,8,9,10]. It typically involves similarity matching with the gene functions predefined in the manually curated databases, such as Gene Ontology (GO)[11], Molecular Signature Database (MSigDB)[12,13] and so on, by rank-based comparison. However, gene sets exhibiting strong enrichment in the existing databases have often been well-validated by previous research, thus an increasing number of studies are shifting their focus towards gene sets that marginally overlap with the known enrichment functions[14]. Their objective is to find worthy biological functions from less-studied cases in GSEA and add the new biological function to the existing databases.

Under this tendency, the powerful reasoning and rich biological context of large language models (LLMs) has drawn the interest of researchers[15,16]. Recent works have utilized instruction learning to prompt LLMs to discover biological mechanisms of gene sets. Hu *et al.*[14] evaluated the performance of five LLMs in gene set analysis. They designed a set of instructions for LLMs to analyze the gene functions and generate a brief biological process name for the given gene set. In their work, GPT-4 demonstrated the highest performance in generating matching name to the ground truth. Using standard LLMs, SPINDOCTOR[17] introduces gene functional synopsis for summarizing and generating multiple biological process names given a gene set. Moreover, the application of GPT-4 in the candidate gene prioritization[18] and genomics question answer[19] also proves the potential of LLMs in gene set knowledge discovery.

However, these studies only employed and evaluated standard large language models (LLMs). Consequently, their results may exhibit common LLM issues such as nondeterministic outputs and uncontrollably inaccurate results (i.e., hallucinations). These shortcomings pose challenges in creating a reliable framework for accurately generating the most prominent biological processes for gene sets and hinder the objective interpretability of gene functions.

In response, we present GeneAgent, a language agent built upon GPT-4 to generate biological process names for gene sets in an interpretable and contextually coherent manner. Such capabilities are directly enabled by autonomously interacting with a variety of biological databases through Web APIs. Utilizing relevant domain-specific information retrieved from expert-curated databases, GeneAgent performs the fact verification, offering objective evidence to support or refute the original output of an LLM. We perform comparative experiments on gene sets from three distinct sources: literature curation (GO); proteomics analysis (NeST system of human cancer proteins[6]); and molecular functions (MSigDB). The evaluation results indicate that GeneAgent significantly outperforms GPT-4 (as previously shown in Hu et al.[14]) in predicting the accuracy of biological processes. Compared with SPINDOCTOR, GeneAgent provides more informative gene functional synopsis for LLMs to generate relevant biological terms. Importantly, GeneAgent achieves such enhanced performance by reducing the occurrences of hallucinations common in standard LLMs.

In a real-world application, we assessed the performance of GeneAgent on seven novel gene sets derived from the mouse B2905 melanoma cell lines. Our findings reveal that GeneAgent not only achieves better performance compared to GPT-4 but also offers valuable insights into novel gene functionalities, facilitating knowledge discovery in the realm of biomedical research. The results of this use case also demonstrates that GeneAgent is robust across different species.

# Results

## GeneAgent Workflow

GeneAgent aims to enhance the accuracy of gene set analysis by minimizing instances of hallucinations, for which we designed a novel feature of autonomous interaction with domain-specific databases, enabling GeneAgent to self-verify and refine the raw output of LLMs (**Method**).

Specifically, the workflow of GeneAgent contains four crucial steps, generation, self-verification, modification, and summarization (**Figure 1a**). GeneAgent creates the process name and analytical texts of gene functions for the input gene set at first. Afterwards, it activates selfVeri-Agent (**Figure 1b**) for the subsequent verification of the process name and analytical texts respectively. Different stages of self-verification are cascaded through the modification module. During each verification, GeneAgent discerns the potential hallucinations by extracting claims from original contents and comparing them against curated knowledge in domain-specific databases. The gene names in claims serve as the basic queries to fetch reference functions from backend databases via Web APIs. Once having the reference functions, selfVeri-Agent will compile the verification report delineating a decision to original claims. Notably, selfVeri-Agent prioritizes the "Process Name" before examining the "Analytical Narratives", ensuring that the process name would be verified twice based on the modified analytical texts. Last, GeneAgent summarizes all intermediate verification reports to produce the final results. Such a cascade structure can improve the traditional step-by-step chain-of-thinking (CoT)[20] and achieve autonomous verification for the inference process[21], as compared with GPT-4 (CoT). In GeneAgent, we utilized domain knowledge curated in 18 biomedical databases via four Web APIs (**Method**).

## GeneAgent significantly outperforms the standard GPT-4.

We compared the accuracy of GeneAgent with GPT-4 proposed by Hu *et al.* in generating the most relevant biological process name for a given gene set. The number of genes in sets ranges from 3 to 456, with an average of 50.67 (**Table 1a**). Please note that we implemented a masking strategy for APIs to ensure no databases is utilized for its own gene sets during the self-verification process (**Method**).

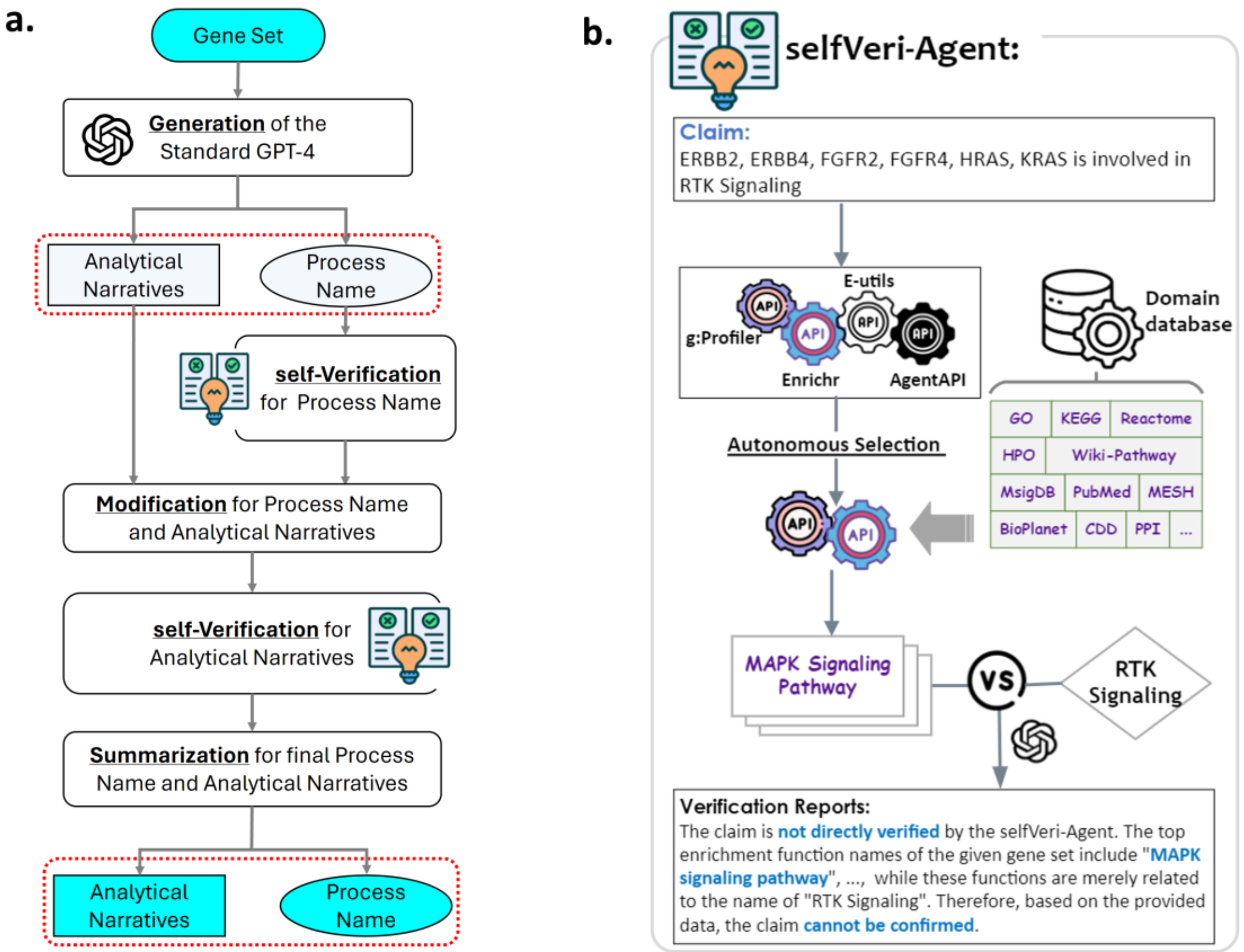

**Figure 1. Framework of GeneAgent for gene set knowledge discovery. a**. The cascade structure of GeneAgent. There are four steps: **G**eneration, self-**V**erification, **M**odification and **S**ummarization. **b**. The workflow of selfVeri-Agent with an example of "ERBB2, ERBB4, FGFR2, FGFR4, HRAS, KRAS is involved in RTK Signaling".

**Table 1. The statistics for gene sets used in our study.**

**Table. 1a, gene sets used for empirical evaluation.**

| Dataset | #sets | #genes | Avg. genes | Avg. words of all reference terms | Resource |
|---|---|---|---|---|---|
| Gene Ontology | 1,000 | 3 to 456 | 48.32 | 4.704 | Literal curation |
| NeST | 50 | 5 to 323 | 18.96 | 2.214 | Proteomics analysis |
| MsigDB | 56 | 4 to 200 | 112.00 | 2.980 | Molecular function |
| All | 1,106 | 3 to 456 | 50.67 | 4.500 | |

**Table 1b, 7 novel gene sets tested in our case study.**

| ID | #genes | Reference term | Resource |
|---|---|---|---|
| mmu05171 (HA-R) | 36 | Coronavirus disease - COVID-19 | Preclinical study of melanoma[29] (Mouse B2905 melanoma cell lines) |
| mmu03010 (HA-R) | 35 | Ribosome | |
| mmu03010 (HA-S) | 49 | Ribosome | |
| mmu05171 (HA-S) | 47 | Coronavirus disease - COVID-19 | |
| mmu04015 (HA-S) | 27 | Rap1 signaling pathway | |
| mmu05100 (HA-S) | 19 | Bacterial invasion of epithelial cells | |
| mmu05022 (LA-S) | 24 | Pathways of neurodegeneration - multiple diseases | |

First, we evaluate ROUGE (Recall-Oriented Understudy for Gisting Evaluation) scores[22] (**Method**) between the generated names and their reference terms. Results (**Figure 2a**) demonstrate that GeneAgent outperforms GPT-4 in generating the same word sequence as reference terms. Among 1,106 gene sets where the average number of words of reference terms is 4.50 (**Table 1a**), GeneAgent achieves much higher scores than that by GPT-4. Notably, GeneAgent improves the Rouge-L (Longest Common Subsequence) and Rouge-1 (1-gram) scores from 23.9% to 31.0% in MsigDB, and Rouge-2 (2-gram) score from 7.4% to 15.5% accordingly.

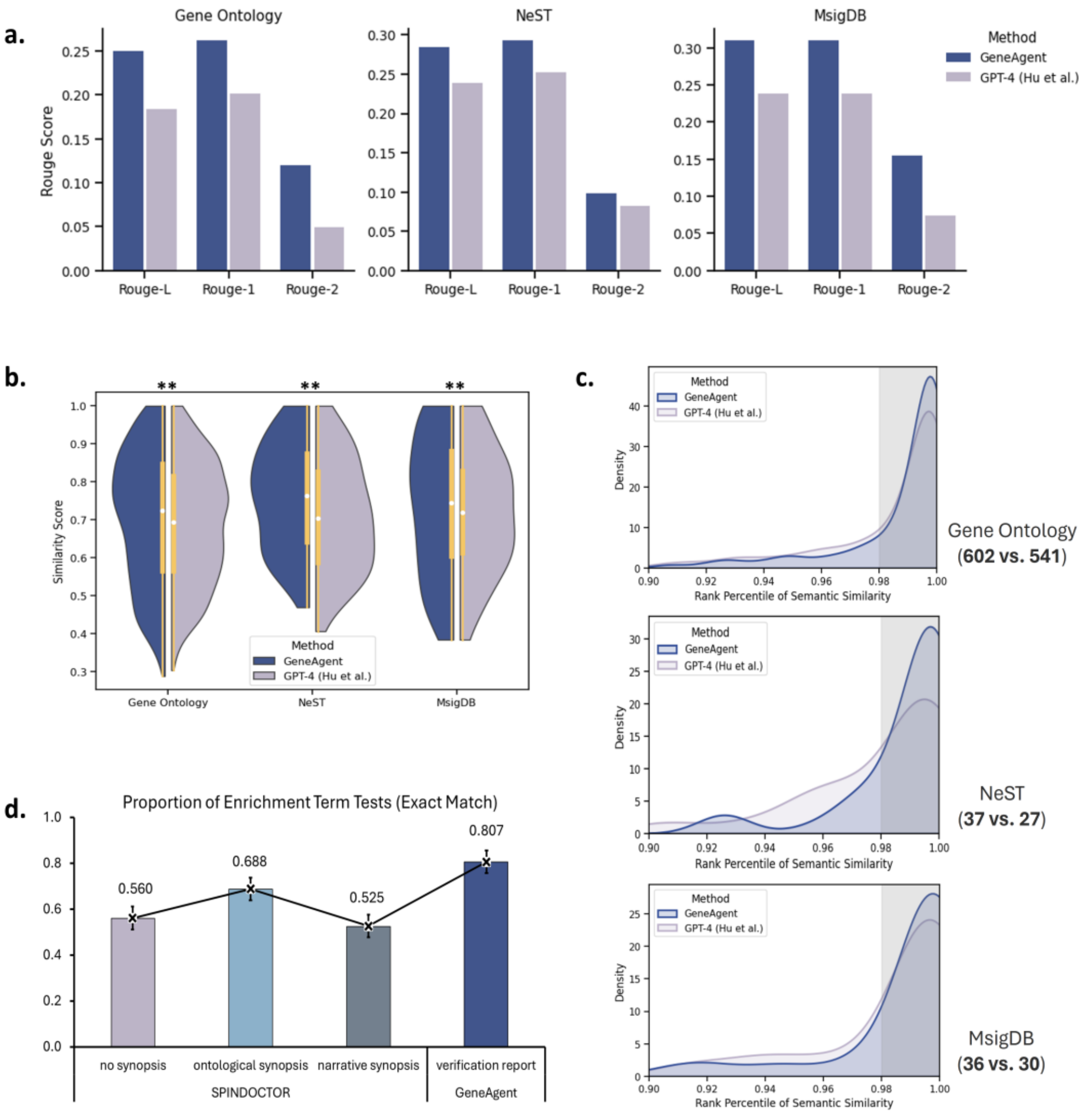

**Figure 2. Compared with GPT-4, GeneAgent generates biological process names for gene sets with a higher similarity to their reference terms. a**. The Rouge score of GeneAgent in three datasets. **b**. Distribution of similarity scores in different dataset. The P-values are calculated by a single-tailed T-test. ** denotes that p-value $< 10^{-3}$. **c**. Distribution of the percentile of semantic similarity between generated names and

their reference terms. The background set contains 12,320 terms consisting of 12,214 GO:BP terms used by Hu *et al*. and all available annotated terms in NeST (50) and MsigDB (56). The plot shown is for the top 90th percentile. The values in each caption denote the number of gene sets in GeneAgent and in GPT-4 at the top 98th percentile. **d.** The proportion of significant enrichment terms in the tested terms based on the exact match.

Next, we measured the semantic similarity between the generated names and their reference terms based on the semantic embeddings encoded by a state-of-the-art biomedical text encoder MedCPT[23] (**Method**). The results (**Figure 2b**) indicate that the average similarity score of GeneAgent is respectively 0.705, 0.761, and 0.736 in three datasets, representing statistically significant improvements (p-value$<$ 0.05) over the GPT-4. Moreover, there are noticeable differences between names generated by GeneAgent and GPT-4 (**Table 2a**). Therefore, we counted the number of gene sets at different levels of similarity sores. GeneAgent generates 170 cases with similarity scores exceeding 90% and 614 cases with similarity scores exceeding 70%. This outcome is much higher than GPT-4 which only has 104 cases and 545 cases. Notably, GeneAgent generates names with a similarity score of exactly 100% in 15 cases, while GPT-4 only generates 3 such cases. A similarity score exceeding 90% indicates the generated name has only subtle differences from its reference term, such as the addition of "Metabolism". A similarity score between 70% and 90% indicates the generated name is a broader concept of the biological process, which would be more similar with the direct ancestor term of the gene set (**Table 2b**).

**GeneAgent generates biological process names that are more related to reference terms than other candidate terms**.

Hu *et al*. introduced the evaluation of "background semantic similarity distribution" in their study, which is estimated by calculating the percentile within a background set of the semantic similarity between the generated name and the reference term. Therefore, we designed the similar pipeline (**Method**) based on MedCPT to evaluate GeneAgent and GPT-4. For example, for the gene set with the term "regulation of cardiac muscle hypertrophy in response to stress", GeneAgent generates the name where the semantic similarity is higher than 98.9% background terms (i.e., at 98.9% percentile) (**Extended Fig. 1a**), while GPT-4 generates the name where semantic similarity is higher than 60.2% background terms (i.e., at 60.2% percentile) (**Extended Fig. 1b**).

**Table 2. Examples of gene sets that are assigned with different biological process names and similarity scores.**

**Table. 2a, gene sets named by different methods.**

| ID | Reference Term | GeneAgent | GPT-4 | GSEA (g:Profiler) |
|---|---|---|---|---|
| GO:0032459 | regulation of protein oligomerization | Protein Sorting and Lipid Transport | Intracellular Protein Transport | Regulation of protein oligomerization |
| NeST:69 | Protein nuclear transport | Nucleocytoplasmic Transport | Telomere Maintenance and Nuclear Transport | protein localization to nucleus |
| MsigDB:69 | Peroxisome | Peroxisome Protein | Peroxisome Biogenesis | protein localization to peroxisome |

**Table 2b, gene sets named by GeneAgent with different similarity scores. Their direct ancestors in GO terms are obtained by g:Profiler.**

| ID | Reference Term | GeneAgent | Similarity Score | Direct ancestor in GO Terms | Similarity with ancestor |
|---|---|---|---|---|---|
| GO:0035108 | limb morphogenesis | Limb Morphogenesis | **1.000** | limb development | 0.928 ↓ |
| GO:0015888 | thiamine transport | Thiamine Transport and Metabolism | **0.989** | vitamin transport | 0.815 ↓ |
| MsigDB:69 | Peroxisome | Peroxisome Protein | **0.957** | peroxisome organization | 0.915 ↓ |
| GO:0048319 | axial mesoderm morphogenesis | Mesodermal Commitment Pathway | 0.772 | mesoderm morphogenesis | **0.829** ↑ |
| NeST:61 | Cullin-RIng ubiquitin ligase complex | Ubiquitin Mediated Proteolysis | 0.826 | ubiquitin ligase complex | **0.910** ↑ |
| NeST:8 | Immune system | Lymphocyte Activation | 0.746 | leukocyte activation | **0.929** ↑ |
| MsigDB:56 | Reactive Oxygen Species Pathway | Response to Oxidative Stress | 0.721 | response to stress | **0.911** ↑ |

For 1,106 gene sets, we presented gene sets whose similarity scores between generated names and reference terms are in the top 90th percentile among 12,320 background terms (**Figure 2c**). Results show that 76.9% (850) of names generated by GeneAgent have a semantic similarity exceeding the 90th percentile (758 from GO, 46 from NeST, and 46 from MsigDB), while GPT-4 yields 742, 42, and 40 gene sets from the respective databases (74.5% in total). In the top 98th percentile, GeneAgent also exhibits a higher performance, with over 675 gene sets surpassing this threshold, compared to 598 for GPT-4. Notably, there are 82 gene sets achieve a 100th percentile in GeneAgent. Conversely, GPT-4 only records 43 instances.

**GeneAgent generates informative synopsis of gene functions for summarizing multiple enrichment terms**.

Inspired by SPINDOCTOR[17], which proposes the summarization of multiple plausible biological process names from the available synopsis of gene functions, we performed the enrichment term test (**Method**) by using GeneAgent's verification report to serve as gene function synopsis. For comparison, we collected the narrative and ontological synopsis of 56 gene sets in MsigDB from the SPINDOCTOR study, and also evaluated the vanilla setting of no gene synopsis.

To assess the enrichment terms summarized from different gene synopsis against those from conventional GSEA, we utilized g:Profiler[24] to extract significant enrichment terms (p-value $\leq$ 0.05) associated with gene sets as the ground truth. Then, we quantified the extent to which generated terms overlapped with the significant terms (**Method**). Our findings, employing an exact match criterion, reveal that 80.7% (296 out of 367) of the LLM-generated terms aligned with significant enrichment terms when using verification reports as the gene synopsis (**Figure 2d**). This proportion declines to 68.8% (282 out of 410) when employing ontological synopsis and further diminishes to 56.0% (195 out of 348) without using gene synopsis. As discussed in SPINDOCTOR, unmatched terms may be instances where the model fabricates a biological function, i.e., hallucination. Therefore, the significantly lower proportion (19.3%) of unmatched terms in GeneAgent underscores its efficacy in mitigating hallucinations.

**GeneAgent mitigates hallucinations by interacting with domain databases**.

Hu *et al*. resort to human inspection to measure the reliability of their GPT-4 pipeline. Conversely, GeneAgent incorporates the proposed self-verification module, acting as an AI agent by autonomously interacting with the domain databases and obtaining relevant knowledge to support or refute raw outputs of an LLM. Consequently, the verification of GeneAgent no longer merely focuses on the response of LLMs but also implements the supervisory of the inference process.

To elucidate its role in our method, we examined 15,903 claims generated by GeneAgent and reported decisions of the selfVeri-Agent. Among these claims, 15,848 (99.6%) were successfully verified, with 84% supported, 1% partially supported, 8% refuted, and remaining 7% unknown (**Figure 3a**). A marginal fraction (0.4%) of claims were not verified due to the absence of gene names necessary for querying pertinent databases through Web APIs.

During the self-verification process, 16% of the claims were not supported. These unsupported claims were distributed across 794 gene sets, representing potential candidates for revision. Of these, 703 (88.5%) were subsequently modified indeed. Furthermore, we analyzed the utilization frequency of Web APIs and their backend databases during the self-verification. The statistic shows a predominant utilization of Enrichr[25,26] and g:Profiler APIs for verifying process names, whereas the validation of analytical texts mainly relies on E-utils[27,28] and CustomAPI (**Figure 3b**). Additionally, GeneAgent interacts with backend databases 19,273 times to verify 15,848 claims (**Figure 3c**), suggesting that each decision is underpinned by evidence retrieved from at least one database. To estimate the accuracy of the self-verification process of GeneAgent, we manually reviewed 10 randomly selected gene sets from NeST with a total of 132 claims, which received 88 supports, 15 partial supports, 28 refutes and 1 unknown by GeneAgent. Human inspections demonstrate that 92% (122) of decisions are correct, indicating a high performance in the self-verification process (**Figure 3d**).

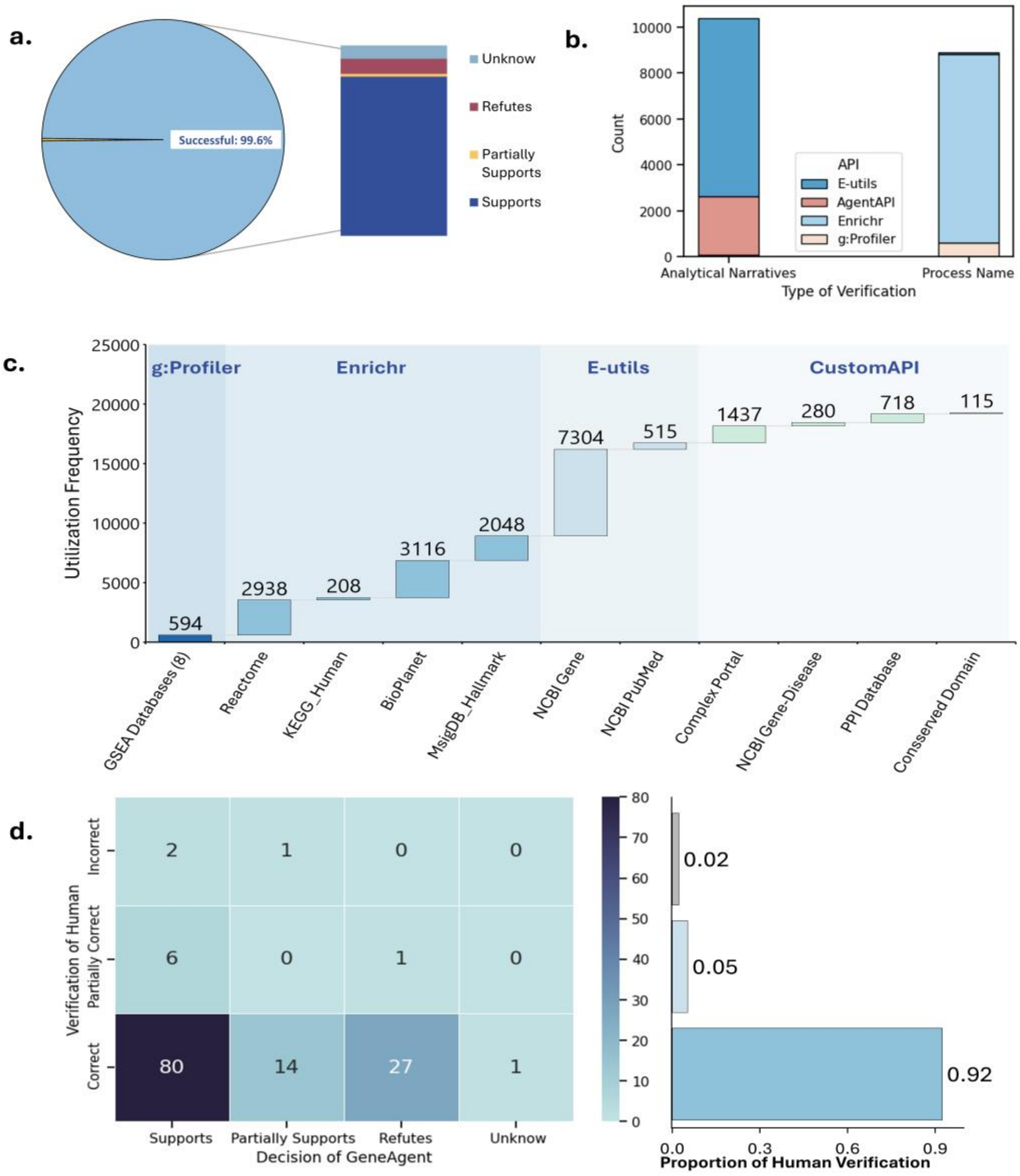

**Figure 3. GeneAgent mitigates hallucinations by autonomously calling Web APIs to interact with domain databases. a**. Statistics of the 15,903 claims collected from the 1,106 gene sets, which contains the proportion of different decisions made by selfVeri-Agent. **b**. Distribution (*y-axis*) of four Web APIs in verifying *Process Name* and *Analytical Texts* (*x-axis*). **c**. The utilization frequency of different backend databases (*x-axis*) in the self-verification stage of GeneAgent. **d**. The results of human verification for the selected 132 claims derived from 10 gene sets.

**GeneAgent offers insightful analytical narratives for novel gene sets.**

As a real-world utilization case, we applied GeneAgent to seven gene sets derived from the study of sub-clonal evolution on gene expression in mouse B2905 melanoma cell lines[29] (**Method**), with the number of genes in each set ranging from 19 to 49 (**Table 1b**).

These gene sets are identified from three subclones to the immunotherapy, i.e., high aggression and resistant (HA-R), high aggression and sensitive (HA-S), and low aggression and sensitive (LA-S). The results (**Table 3**) show that GeneAgent outperforms GPT-4 in generating correct process names and drafting informative analytical narratives.

On the one hand, two gene sets, i.e., mmu04015 (HA-S) and mmu05100 (HA-S), are assigned with process names that exhibit perfect alignment with established reference terms by domain experts. On the other hand, GeneAgent reveals novel biological insights for specific genes in the gene set. Taking mmu05022 (LA-S) for instance, GeneAgent suggests gene functions related to subunits of Complex I, IV, and V in the mitochondrial respiratory chain complexes[30], and further summarizes the "Respiratory Chain Complex" for these genes (**Extended Fig. 2a**). However, GPT-4 categorizes these genes into the "Oxidative Phosphorylation", which is a high-level biological process based on the mitochondrial respiratory chain complexes[31,32], without including the gene Ndufa10 into this process. Similarly, GPT-4 does not include the gene Atxn1l into "Neurodegeneration" and does not provide biological function of the gene Gpx7 (**Extended Fig. 2b**). Such results suggest that GeneAgent is more robust than GPT-4 on novel gene sets, and that GeneAgent is applicable to non-human genes.

To further measure the quality of outputs generated by GeneAgent and GPT-4, we formulated four criteria that are recognized as critical in practical uses by genomic researchers: Relevance, Readability, Consistency, and Comprehensiveness (**Method**). Two experts were recruited to manually assess and compare results (**Table 3**). In terms of readability and consistency, GeneAgent and GPT-4 both demonstrate excellence across numerous cases. But with regards to relevance and comprehensiveness, GeneAgent outperforms GPT-4, which can be attributed to its access to domain-specific databases during the verification stage, thereby offering potentially valuable insights for experts. Nonetheless, there is one case, i.e., mmu03010 (HA-S), where neither GeneAgent nor GPT-4 produces satisfactory results based on the four criteria. GeneAgent generates a narrow process name "cytosolic ribosomes" that does not cover mitochondrial ribosomal genes such as Mrpl10 and Mrps21, while GPT-4 generates a hallucinated response "Synthesis".

**Table 3. Human annotation for the output of GeneAgent and GPT-4 in the case study. “*GPT-4*” is the abbreviation of GPT-4. “○” denotes the better one in each criterion. “✓” denotes the better one in final decision. “×” denotes unreasonableness in output. “Blank cells” denotes both perform well.**

| ID | Generated by GPT-4 | Generated by GeneAgent | Gene Coverage in the Output | Better Output Annotated by Genomic Experts | | | | | | | | | |
|---|---|---|---|---|---|---|---|---|---|---|---|---|---|
| | | | | Relevance | | Readability | | Consistency | | Comprehensive | | Final Decision | |
| | | | | GPT-4 | GeneAgent | GPT-4 | GeneAgent | GPT-4 | GeneAgent | GPT-4 | GeneAgent | GPT-4 | GeneAgent |
| mmu05171 (HA-R) | Ribosomal Protein Synthesis | Cytosolic Ribosome and Protein Synthesis | 33/36 | | ○ | | | | | | | | ✓ |
| mmu03010 (HA-R) | Ribosomal Protein Synthesis and Assembly | Cytosolic Ribosome | 34/35 | | ○ | | | | | | | | ✓ |
| mmu03010 (HA-S) | Ribosomal Protein Synthesis | Cytosolic Ribosome | 13/49 | | | | | | | | | × | × |
| mmu05171 (HA-S) | Ribosomal Protein Synthesis | Cytosolic Ribosome Assembly and Protein Synthesis | 47/47 | | ○ | | | | ○ | | ○ | | ✓ |
| mmu04015 (HA-S) | MAPK/ERK Pathway Regulation | Rap1 Signaling Pathway | 27/27 | | ○ | | | | | | | | ✓ |
| mmu05100 (HA-S) | Caveolae-Mediated Endocytosis and Actin Remodeling | Bacterial Invasion of Epithelial Cells | 19/19 | ○ | | | | ○ | | ○ | | ✓ | |
| mmu05022 (LA-S) | Oxidative Phosphorylation and Neurodegeneration | Neurodegeneration and Respiratory Chain Complex | 23/24 | | ○ | ○ | | | | | ○ | | ✓ |

## Discussion

**Self-verification in GeneAgent**. Recent research has increasingly focused on the "self-verification" strategy within LLMs[33,34,35]. These studies utilized the same LLM to verify its own outputs, which may also lead to overconfidence in the raw results, and potentially heightens the risk of failing to discover novel insights, as the models might not adequately question or critique their initial findings[36]. Differently, GeneAgent leverages established knowledge from manually curated domain-specific databases to verify the initial outputs (**Figure 1b**), which can not only mitigate the overconfidence in the initial results, but aids in reducing potential hallucinations and enhancing the reliability of LLMs.

**GeneAgent versus GSEA**. As an indispensable tool for gene set analysis, GSEA produces the most informative term for gene sets along with the statistical information. In GeneAgent, we included four different APIs (e.g., g:Profiler) to ascertain the agreement of gene sets with those represented by expert-curated databases. Through the comparison between generated names and the most significant enrichment term produced by the g:Profiler tool, we found that GeneAgent surpasses GESA in terms of both similarity (**Extended Fig. 3a**) and ROUGE scores (**Extended Fig. 3b**). In addition to superior performance, GeneAgent can generate associated narratives, which increases the transparency of AI results and explains the biological roles of genes. Therefore, GeneAgent can essentially be seen as a system that merges the strengths of both LLMs and GSEA, delivering performance that surpasses each individual system.

**Importance of expert-curated domain databases.** In addition to the eight databases utilized in the GSEA tool, we have incorporated four databases for pathway analysis and six for gene functional verification (**Figure 3c**). These databases formed a cohesive system that facilitates the discovery of gene set knowledge by providing a reliable foundation of gene functions. The databases used in GSEA are complemented by the others, especially for examining the consistency of individual genes and their shared functions. This is particularly vital for uncovering latent biological functions among multiple genes, as it offers detailed insights into the characteristics of individual genes. Taken together, the domain-specific databases curated by experts are essential for enhancing the effectiveness of GeneAgent in the discovery of gene set knowledge.

**Error analysis**. We analyzed three representative gene sets that received low similarity scores across the three datasets, along with their analytical narratives and verification reports (**Extended Tab. 1**). The suboptimal performance of GeneAgent in those cases can be primarily attributed to two factors: the erroneous rejection of an accurate process name, such as “EGFR Signaling Pathway Regulation” and “Prostate Cancer Progression”; and the incorrect endorsement of an originally dissimilar process name, exemplified by “Catecholamine Biosynthesis”. Employing additional relevant domain databases in the self-verification stage or engineering a more effectiveness prompts in the modification stage may help alleviate such issues.

**Limitations** In this work we only selected GPT-4 as the backbone model given its superior performance. While other LLMs can also be explored, Hu et al., shows that GPT-4 outperforms GPT-3.5, Gemini-Pro, Mixtral-Instruct and Llama 2. While the self-verification step is shown to be effective, GeneAgent might still generate the biological process names that are highly different from their reference terms, which can be potentially alleviated by employing more relevant domain databases in future works. Finally, our study has not attempted to pre-process the gene set such as removing the genes that are non-coherent with other genes from a gene set. Nonetheless, GeneAgent demonstrates remarkable robustness across various gene sets and different species, and effectively mitigates hallucinations by automatically interacting with domain-specific databases.

# Online Method

**Overview of GeneAgent**.

GeneAgent is an AI agent composed of four key modules: generation, self-verification, modification, and summarization. Each module is triggered by a specific instruction tailored to its function. The goal of GeneAgent is to generate a representative biological process name ($P$) for a set of genes, denoted as $D = \{g_i|_{i=1}^{N}\}$. Each gene $g_i$ is this set is identified by its unique name, and the $D$ is associated with a specific reference biological term ($G$). When provided with an $D$, GeneAgent outputs an $P$, accompanied by analytical texts ($A$) detailing the functions of the genes involved, which can be formally defined as $GeneAgent(D) = (P, A)$. In our research, we utilized the GPT-4 model (version 20230613 via the Azure OpenAI API) with the temperature parameter set to 0, ensuring consistent and stable output.

**Pipeline of generating the most prominent biological process names for gene sets**.

The gene set in the dataset $D$ is separated by commas (",") and serves as input parameters for the instruction of the generation ($g$) module. Following the generation stage, $D$ is assigned an initial process name ($P_{ini}$) and corresponding analytical narratives ($A_{ini}$), i.e., $GeneAgent_g(D) = (P_{ini}, A_{ini})$.

Afterwards, GeneAgent generates a list of claims for $P_{ini}$ by using statements like "be involved in" and "related to" to generate hypothesis for gene set and its process name. After that, GeneAgent activates selfVeri-Agent (**Figure 1b**) to verify each claim in the list. Initially, selfVeri-Agent extracts all gene names and the process name from the claims. Subsequently, it utilizes the gene names to invoke the appropriate APIs for the autonomous interaction with domain-specific databases, employing their established knowledge to validate the accuracy of the process name. Finally, it assembles a verification report ($\mathcal{R}_P$) that contains findings and decisions related to the original claim.

Next, GeneAgent initiates the modification ($m$) stage to either revise or retain the $P_{ini}$ based on the findings in the $\mathcal{R}_P$. If the $P_{ini}$ is determined to revise by GeneAgent, the $A_{ini}$ is also instructed to be modified accordingly, i.e., $GeneAgent_m(P_{ini}, A_{ini}, \mathcal{R}_P) = (P_{mod}, A_{mod})$. Following this, GeneAgent applies the self-verification to the $A_{mod}$ to verify the gene functions in the analytical narratives while checking the new process name again. This step is also started with generating a list of claims for different gene names and their functional terms and is finished with deriving a new verification report ($\mathcal{R}_A$) by selfVeri-Agent.

Finally, based on the report $\mathcal{R}_A$, both $P_{mod}$ and $A_{mod}$ are modified according to the summarization ($s$) instruction to generate the final biological process name ($P$) and the analytical narratives ($A$) of gene functions, i.e., $GeneAgent_s\ (P_{mod}, A_{mod}, \mathcal{R}_A) = (P, A)$.

**Domain-specific databases configured in selfVeri-Agent.**

In the self-verification stage, we have configured four Web APIs to access 18 domain databases (**Figure 3c**).

**g:Profiler**[24] (https://biit.cs.ut.ee/gprofiler/page/apis) is an open-source tool for GSEA. In GeneAgent, we used eight domain-specific databases such as GO, KEGG[37], Reactome[38], WikiPathways[39], Transfac[40], miRTarBase[41], CORUM protein complexes[42], and Human Phenotype Ontology[43] to perform enrichment analysis for the gene set. For each gene set, we employed g:GOSt interface to identify top-5 enrichment terms along with their descriptions.

**Enrichr**[25,26] (https://maayanlab.cloud/Enrichr/help\#api) is also a valuable tool for GSEA. We configured four databases related to the pathway analysis in the Enrichr API, i.e., KEGG_2021_Human, Reactome_2022, BioPlanet_2019[44], and MsigDB_Hallmark_2020. In GeneAgent, we selected to return the top-5 standard pathway names via databases.

**E-utils**[27,28] (https://www.ncbi.nlm.nih.gov/) is an API designed for accessing the NCBI databases for various biological data. In GeneAgent, we augment our repository of functional information associated with an individual gene by invoking its Gene database and PubMed database. Different databases can be used by defining the db parameter as gene or pubmed in the basic API.

**CustomAPI** is our custom API library, developed using four gene-centric databases related to gene-disease, gene-domain, PPI, and gene-complex. In GeneAgent, we invoke the appropriate database by specifying the desired interface at the end of the basic API, and subsequently retrieving the top-10 relevant IDs to gene functions. These IDs are then used to match their names in the corresponding database.

Notably, we implemented a masking strategy for APIs and databases during the evaluation to ensure unbiased assessments across various gene sets. Specifically, we removed the g:Profiler API when assessing gene sets from the Gene Ontology dataset because it can perfectly derive their reference terms. Similarly, we masked the "MsigDB_Hallmark_2020" database within the Enrichr API when evaluating gene sets from MsigDB.

**Calculation of ROUGE score**.

Three distinct Rouge metrics[22] are employed to access the recall of generated names relative to reference terms: i.e., Rouge-1 and Rouge-2, which based on n-gram, and Rouge-L, which utilizes the longest common subsequence (LCS). The calculation formulas are as follows:

$$\text{Rouge-N} = \frac{\sum_{S \in ref} \sum_{g_N \in S} count_{match}(g_N)}{\sum_{S \in ref} \sum_{g_N \in S} count(g_N)}, \ \text{N} = 1, 2 \quad (1)$$

$$\begin{cases} R_{lcs} = \frac{LCS(ref, hyp)}{m} \\ P_{lcs} = \frac{LCS(ref, hyp)}{n} \\ \text{Rouge-L} = \frac{(1+\beta^2) R_{lcs} P_{lcs}}{R_{lcs} + \beta^2 P_{lcs}} \end{cases} \quad (2)$$

Here, the $ref$ denotes the reference terms and $hyp$ denotes the generated names. $m$ and $n$ is the token length of $ref$ and $hyp$ respectively. $\beta$ is a hyper-parameter.

**Calculation of semantic similarity**.

After generating biological process name $(P)$ for the gene set $D$, the semantic similarity between $P$ and its reference term $(G)$ is computed by MedCPT[23], a state-of-the-art model for language representation in the biomedical domain. It is built based on PubMedBERT[45] with further training using 255 million query-article pairs from PubMed search logs. Compared with SapBERT[46], BioBERT[47], etc., MedCPT has higher performance in encoding the semantics of biomedical texts.

a) Calculation of semantic similarity between $P$ and $G$

First, $P$ and $G$ are encoded by MedCPT into embeddings, and then the cosine similarity between their two embeddings is calculated, yielding a score in the interval [-1, 1]. Finally, we take the average value of all similarity scores to evaluate the performance of GeneAgent on gene sets in one dataset.

b) Calculation of background semantic similarity distribution

First, $P$ is paired with all possible terms $G_i \in Q$, where $Q$ denotes 12,320 background terms consisting of 12,214 GO:BP terms in GO, and all available terms in NeST (50) and MsigDB (56). Then, $P$ and $G_i$ are fed into MedCPT to get the embeddings, i.e., $\vec{P}$ and $\vec{G_i}$. Afterwards, we calculated the cosine similarity for each $< \vec{P}, \vec{G_i} >$ pair. Finally, we ranked all cosine scores from large to small and observed the position where the pair $< P, G_p >$ ($G_p$ is the reference term for $P$) located in. The higher position denotes the generated names have a higher similarity score to their reference terms than other terms.

**Pipeline of enrichment term test based on verification reports of GeneAgent**.

For gene sets in MsigDB, we first collected its verification report produced by GeneAgent. Afterwards, each gene set and the associated report were used as the parameters of the instruction for the GPT-4. Therefore, GPT-4 can summarize multiple enrichment terms for the given gene set. Finally, we employed the exact match to evaluate the accuracy of the tested terms summarized by the GPT-4. Specifically, for each gene set in the evaluation, we first utilized g:Profiler to perform GSEA, where the p-value threshold is set to 0.05. Then, we obtained significant enrichment terms for the given gene sets as the ground truth. Finally, we counted the number of tested terms summarized by GPT-4 that correctly match the significant enrichment term of each gene set. One tested term is deemed accurate only when all words are exactly matched with all words of one term in the ground truth. One tested term is considered as accurate only when there is an exact match between all the words in the tested term and one term in the ground truth.

**Human checking for the decisions in the verification report of GeneAgent**.

We randomly selected 10 gene sets from NeST with 132 claims for human inspection. There are two parts in the verification report: the claims and the decisions to the claims along with evidence. Annotators were asked to label the selfVeri-Agent decisions (i.e., supports, partial supports, and refutes) for each claim and judge whether such decisions are correct, partially correct, or incorrect, which follows the study of natural language inference[48] and fact verification[49]. For each claim, the annotators need to make a judgment based on assertions of the gene (set) functions provided in the evidence:

a) **Correct**: This category applies when GeneAgent's decision completely aligns with the evidence supporting the original claim. The decision is considered correct if it accurately reflects the evidence documented, demonstrating a clear and direct connection between the claim and the supporting data.

b) **Partially correct**: It is designated when GeneAgent's decision requires indirect reasoning or when the decision, although related, does not completely align with the direct evidence provided. This occurs when the decision is somewhat supported by the evidence but requires additional inference or context to be fully understood as supporting the original claim.

c) **Incorrect**: This category is used when GeneAgent's decision either contradicts the evidence or lacks any substantiation from the verification report. It includes decisions that misinterpret the evidence or ignore significant aspects of the evidence.

**Melanomas gene sets in the preclinical study.**

The mouse B2905 melanoma cell line, which is derived from a tumor from the M4 model, where melanoma is induced by UV irradiation on pups of hepatocyte growth factor (HGF)-transgenic C57BL/6 mice[50]. Specifically, 24 single cells were isolated from the parental B2905 melanoma line and then expanded to become individual clonal sublines (i.e., C1 to C24)[51]. Each of these 24 sublines was subjected to whole exome sequencing and full-transcript single-cell RNA (scRNA) sequencing by Smart-seq2 protocol. The single nucleotide variants called from exome sequencing results were used to build the tumor progression tree for all the 24 sublines. Based on the in vivo growth and therapeutic responses of the sublines in the clusters, three clades are named as "high aggressiveness and resistant (HA-R)", "high aggressiveness and sensitive (HA-S)", and "low aggressiveness and sensitive (LA-S)"[29]. Afterwards, EvoGeneX[52] is applied to the scRNA data of the 24 clonal sublines, where the phylogenetic relation is defined by the mutation-based tumor progression tree, to identify adaptively up-regulated and down-regulated genes in each of HA-R, HA-S, and LA-S clades. The adaptively up- and down-regulated gene lists were then subjected to the KEGG pathway enrichment analysis. The genes in the enrichments and their enriched terms are used to test the GeneAgent.

**Human annotation for outputs in the case study.**

For the assessment of different outputs between GeneAgent and GPT-4, we established four criteria following the existing studies on the evaluation of LLM[53,54].

a) **Relevance**: Assess whether the content about genes pertinently reflects their functions, providing value to biologists.

b) **Readability**: Evaluate the fluency and clarity of the writing, ensuring it is easily understandable.

c) **Consistency**: Determine whether the analytical narratives align consistently with the specified process name.

d) **Comprehensiveness**: Verify whether the outputs provide a comprehensive understanding of gene functions.

Based on these four established criteria, two experts are tasked with evaluating the final responses from the outputs of GPT-4 and GeneAgent. They operate the annotation under a blind assessment protocol, where they are unaware of the algorithm that produced each response. Their main responsibility is to annotate and compare the preference for outputs generated by GPT-4 versus GeneAgent. They carefully review and select the more effective response, justifying their selections with relevant comments. Following a comprehensive synthesis of all feedback, these two experts are required to make a definitive judgment on which output most effectively satisfies the users' requirement.

# Extended Data Figures and Tables

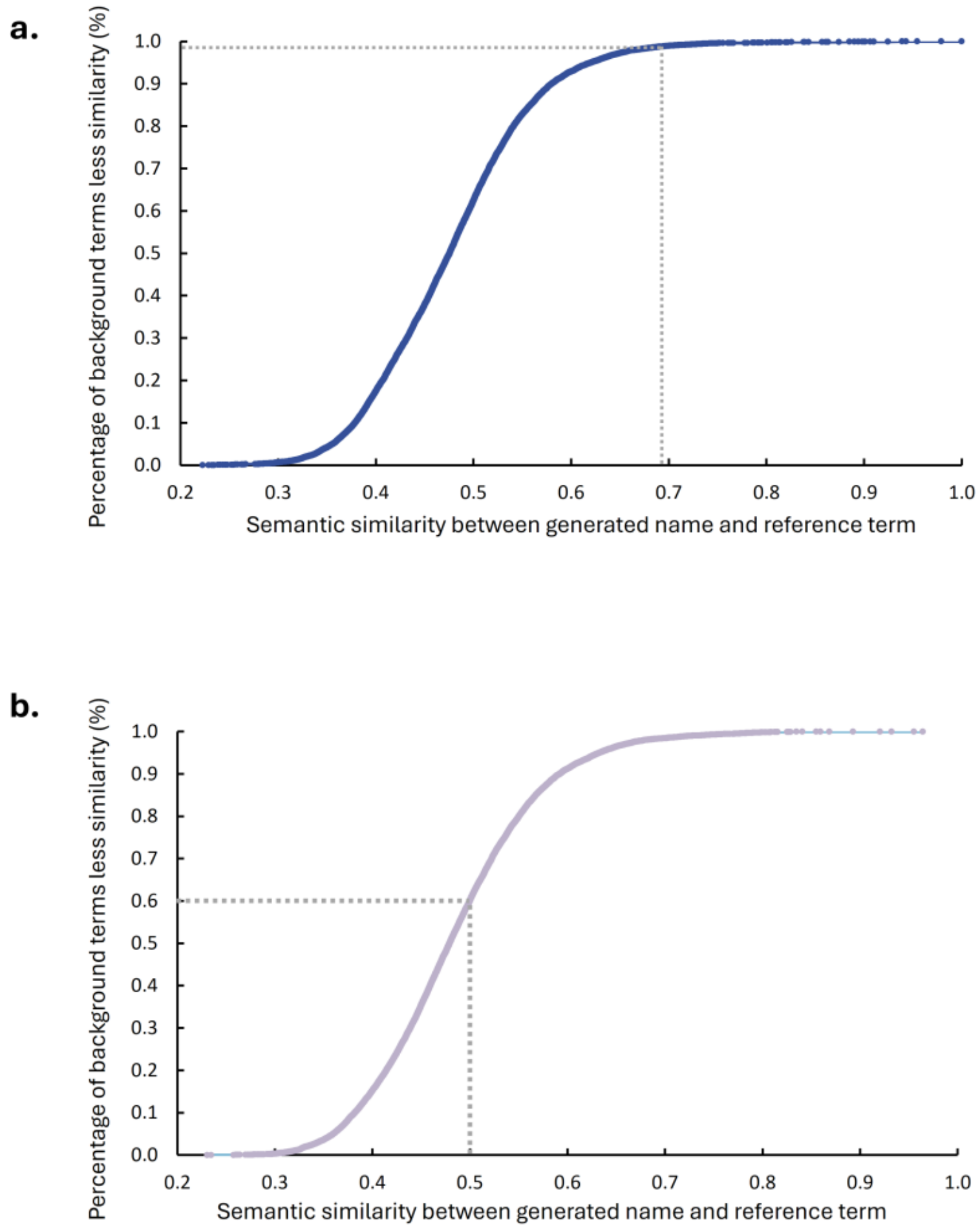

**Extended Fig. 1. Semantic similarity between generated name and reference term (*gray dashed line, x-axis*) is converted to the percentage of all terms in the background set with lower similarity to the generated name (*gray dashed line, y-axis*). a**. Example of the reference term ("*regulation of cardiac muscle hypertrophy in response to stress"*) and the generated name of GeneAgent ("*Regulation of Cellular Response to Stress"*). The similarity of reference term and generated name is 0.695, which is higher than other 98.9% terms in the background set. **b.** Example of the reference term ("*regulation of cardiac muscle hypertrophy in response to stress"*) and the generated name of GPT-4 ("*Calcium Signaling Pathway Regulation"*). The similarity of reference term and generated name is 0.500, which is higher than other 60.2% terms in the background set.

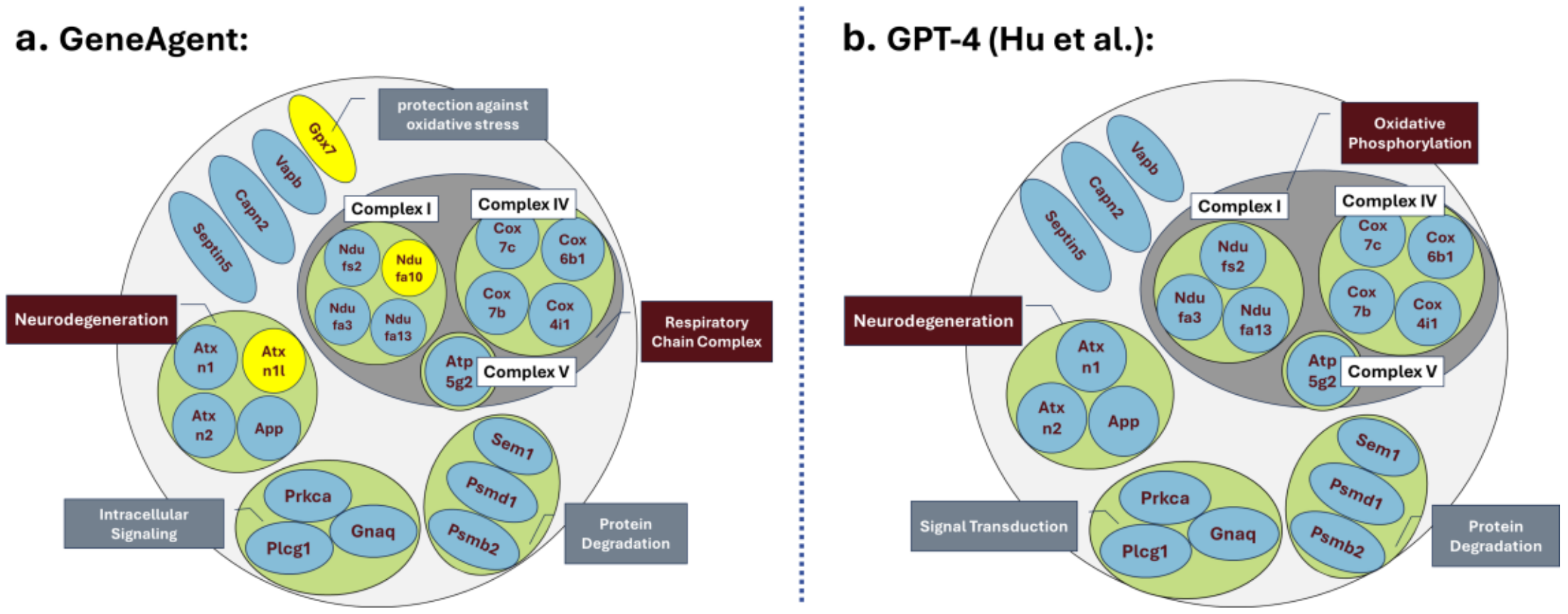

**Extended Fig. 2. Example of gene functions concluded by GeneAgent and GPT-4.** The example shown for the gene set "mmu05022 (LA-S)" in the case study. **a**. GeneAgent takes the "Neurodegeneration and Respiratory Chain Complex" as the most prominent biological process. Complex I is the ubiquinone oxidoreductase, Complex IV is the cytochrome c oxidase and Complex V is the ATP synthase. **b.** GPT-4 (Hu et al.) takes the "Oxidative Phosphorylation and Neurodegeneration" as the most prominent biological process.

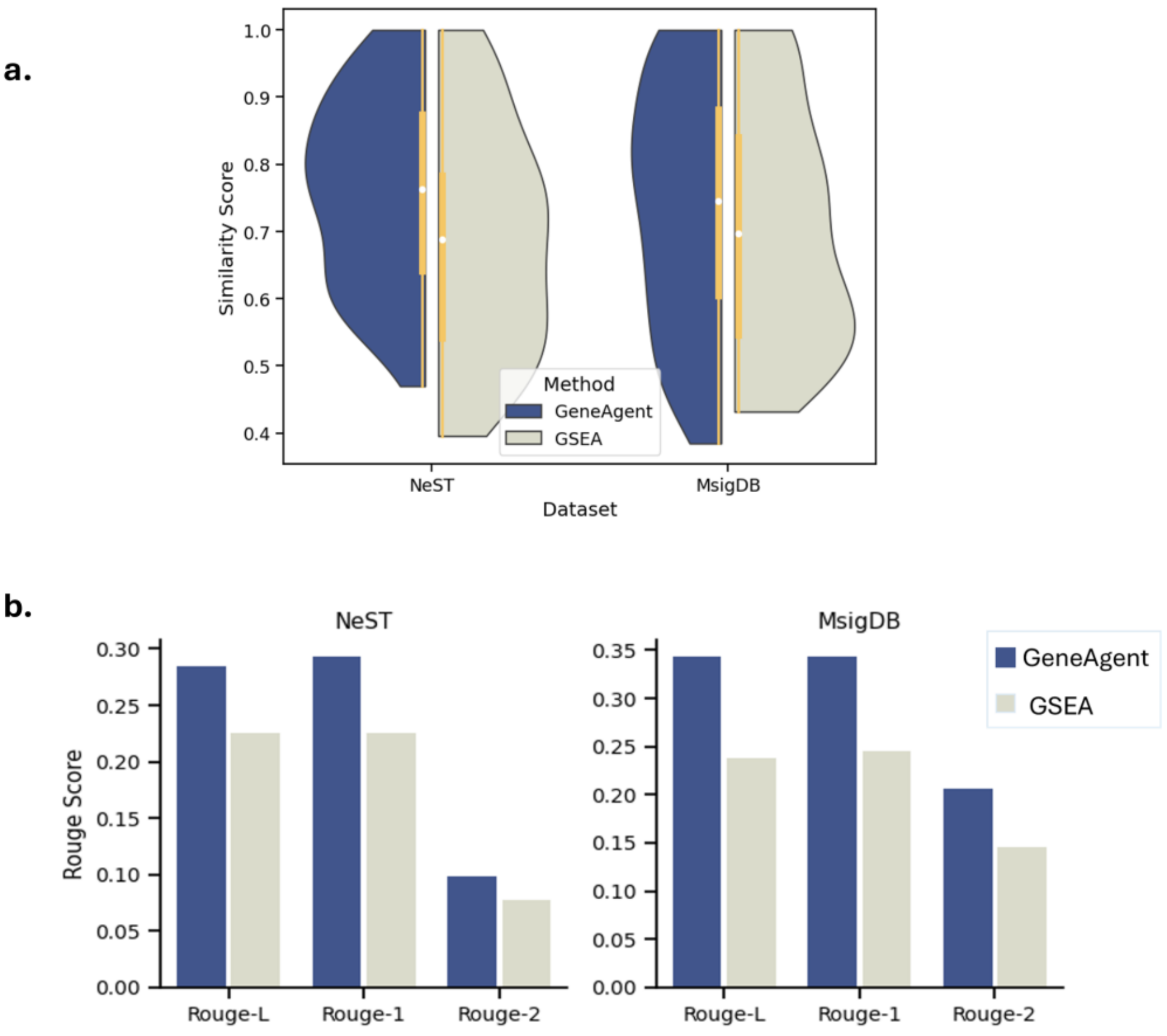

**Extended Fig. 3. Complementary experiments between the GeneAgent and the conventional GSEA (g:Profiler). a**. Comparison of semantic similarity. **b**. Comparison of Rouge scores.

**Extended Tab. 1. Gene Sets with low semantic similarity score.** **XXX (**)** denotes the name generated by GPT-4 (Hu *et al.*) and the semantic similarity to the reference term. **XXX** denotes the enrichment results returned by the domain databases.

| ID | Reference Term | Name generated by GeneAgent | Similarity Score | Major Evidence in the Verification Report | Root Causes of Poor Performance |
|---|---|---|---|---|---|
| NEST:169 | Neg Regulation EGFR | Cell Proliferation and Adhesion Regulation | 0.470 | The claim that the **EGFR Signaling Pathway Regulation (0.739)** is regulated by the gene set AKT1, CTNNB1, EGF, EGFR, NF2, PTEN is not directly supported by the data. The top-5 enrichment function names of the regulation, signaling pathway, and complex for the given gene set include **Endometrial cancer, Prostate cancer, Embryonic stem cell pluripotency pathways, and Breast cancer**. | **Incorrectly refute original Process Name**: The original similar Process Name generated by standard GPT-4 is refuted. |
| MsigDB:12 | Androgen Response | Cytoplasmic Protein Interaction and Regulation | 0.384 | [1]. The gene set provided is indeed associated with **prostate cancer progression (0.615).** The top-5 enrichment function names for this gene set include **"cytoplasm", "prostate; glandular cells [High]", "prostate; glandular cells [≥Medium]", "extracellular exosome", and "extracellular vesicle"**.<br>[2]. The claim that the gene CDK6 is associated with the progression of prostate cancer cannot be verified.<br>[3]. The claim that KLK2 and KLK3 genes are well-known biomarkers for prostate cancer cannot be fully verified. | **Incorrectly refute original Process Name**: The original similar Process Name generated by standard GPT-4 is supported but the is refuted by the verification for genes in the Analytical Narratives. |
| GO:0046684 | response to pyrethroid | Catecholamine Biosynthesis | 0.369 | The claim that the process of **Catecholamine Biosynthesis (0.369)** is facilitated by the human gene set DDC, TH, SCN2B is supported. The gene set DDC, TH is involved in several biological pathways related to **neurotransmitter disorders, dopamine metabolism, biogenic amine biosynthesis, and amine-derived hormones**. However, the gene SCN2B does not appear to be involved in these pathways. | **Incorrectly support the original Process Name**: The original dissimilar Process name generated by standard GPT-4 is supported. |

## Data Availability

Publicly available gene sets were used in this study. Gene Ontology (2023-11-15 release) and the selected NeST gene sets used in the study of Hu et al. are available at https://github.com/idekerlab/llm_evaluation_for_gene_set_interpretation/blob/main/data/. Gene sets used in the MsigDB dataset are the subset of data used in the research of https://github.com/monarch-initiative/talisman-paper/tree/main/genesets/human.

## Author contributions

**Z.W.**, **Q.J.**, and **Z.L.** conceived this study. **Z.W.** and **Q.J.** implemented the data collection and model construction. **Z.W.** conducted model evaluation and manuscript drafting. **C.W.**, **S.T.**, and **P.L.** developed the CustomAPI Library. **S.T.**, **C.W.** and **Z.W.** developed the demo website of GeneAgent. **C.D.** provided the gene sets derived from the mouse B2905 melanoma cell line. **Z.W.** and **Q.Z**. contributed to the data annotation in the self-verification. **C.D.** and **C.R.** contributed to the data annotation in the case study. **Z.L.** supervised the study. All authors contributed to writing the manuscript and approved the submitted version.

## Acknowledgements

We would like to thank Xiuying Chen, M.G. Hirsch, and Teresa M. Przytycka for their helpful discussion of this work. This research is supported by the NIH Intramural Research Program, National Library of Medicine.